# Blind Image Quality Assessment Using Multi-Stream Architecture with Spatial and Channel Attention

Muhammad Azeem Aslam1,*, Xu Wei2, Hassan Khalid3, Nisar Ahmed4, Zhu Shuangtong2, Xin Liu1, and Yimei Xu1
1School of Information Engineering, Xi'an Eurasia University, Xi'an, 710065, Shaanxi, China.
2Changchun Institute of Optics, Fine Mechanics and Physics, Chinese Academy of Sciences, Changchun 130033, China.
3Department of Electrical Engineering, University of Engineering and Technology Lahore, Lahore, 54890, Punjab, Pakistan.
4Department of Computer Engineering, University of Engineering and Technology Lahore, Lahore, 54890, Punjab, Pakistan.
*Email: azeem@eurasia.edu

**Abstract:**

BIQA (Blind Image Quality Assessment) is an important field of study that evaluates images automatically. Although significant progress has been made, blind image quality assessment remains a difficult task since images vary in content and distortions. Most algorithms generate quality without emphasizing the important region of interest. In order to solve this, a multi-stream spatial and channel attention-based algorithm is being proposed. This algorithm generates more accurate predictions with a high correlation to human perceptual assessment by combining hybrid features from two different backbones, followed by spatial and channel attention to provide high weights to the region of interest. Four legacy image quality assessment datasets are used to validate the effectiveness of our proposed approach. Authentic and synthetic distortion image databases are used to demonstrate the effectiveness of the proposed method, and we show that it has excellent generalization properties with a particular focus on the perceptual foreground information.

## 1 Introduction

Image Quality Assessment (IQA) is an important research problem in the domain of computer vision, and the importance has been increased with the evolution of digital multimedia technologies. Digital images are widely used in various fields, including medicine, surveillance, remote sensing, and entertainment. These images undergo various type of deterioration which are introduced during the process of acquisition, compression, storage or communication. Therefore, it is pragmatic to check the quality of the images to ensure they be used for their intended purpose efficiently.

Digital images are affected by various artifacts and distortions during the process of acquisition, compression, transmission, and post-processing. Moreover, factors such as imaging sensor's limitations, transmission bandwidth, and storage space, introduce artifacts such as noise, blur, compression artifacts, and distorted colors. These distortions can reduce image quality and make it difficult to see important details. These modifications in the original information can have a significant impact on the image's quality. For example, in the medical field, distorted images can lead to inappropriate diagnosis and so treatment. Similarly, bad image quality can make it difficult for people to enjoy what they are watching when it comes to entertainment and so can significantly decrease the Quality of Service (QoS). Distortions have varying effects on image quality depending on the application. For example, low-quality images may be useful in surveillance applications for detecting specific features, but even minor distortions in medical imaging can have serious consequences. The introduction of artifacts and distortions into digital images can have a significant impact on their quality, and understanding the effect of these distortions is critical in order to mitigate their effect and so a generalized image quality assessment algorithm is required [1-5].

Image quality assessment (IQA) algorithms can be broadly categorized into subjective and objective IQA algorithms. Subjective IQA involves human evaluators who rate the quality of images and afterwards a mean score for each image has been calculated, the same is then used as the quality score for the image. This method is considered as the most accurate and ideal, but it is time-consuming, costly, and not suitable for real-time applications. Objective IQA uses algorithms that emulates the human perception of quality, and so making them suitable for real-time applications. Objective IQA algorithms can be further divided into three types: full reference, reduced reference, and No-reference IQA. Full reference IQA algorithms compare the original and distorted images, while reduced reference IQA algorithms use metadata from the original image to determine the quality of the distorted image. No-Reference IQA algorithms determine the quality of the distorted image without any information of the original or pristine image. Each type of IQA algorithms have their own advantages and limitations, and the choice of algorithm depends on the specific application and requirements. To put in nut shell subjective IQA provides the most accurate results, while objective IQA algorithms can be used effectively in real-time applications. Full reference, reduced reference, and no-reference IQA algorithms provide different trade-offs between accuracy and practicality. Among, all the methods the no-reference or blind image quality assessment algorithms have gain significant attention in the last decade with the evolution of social media and exponential flow of image data. The relationship between different techniques for IQA algorithms is reported in Figure 1 [1-22].

The blind image quality assessment (BIQA) is a challenging problem that aims to evaluate the quality of an image without any reference information. There are two major categories of BIQA approaches: handcrafted feature-based and deep learning-based.

Handcrafted feature-based methods take a set of features from an image and use them to predict its quality in integration with machine learning algorithms. Mostly, these features are designed to capture some specific aspects of the useful information from the input image, like sharpness, contrast, statistical information, and color. The generated features are very useful; however, these features may still have some limitations. For example, they may not be able to capture complex distortions such as: fade, noise, blur and other artifacts which occurs simultaneously [5, 20-25].

The deep learning-based methods have shown state-of-the-art performance in BIQA problem. Convolutional neural networks (CNNs) are basically used in these methods to automatically learn a set of features that are best for predicting the quality of an image. CNNs can capture complex information about the distortions and can efficiently learn features that are more robust than handcrafted features [1-4, 6-13, 19].

For scientific rigor, it is important to use the right metrics and datasets to evaluate how well a BIQA algorithm work. Therefore, for the evaluation of the performance of BIQA algorithm in a systematic way, several metrics have been proposed in literature. The most commonly used metrics for correlation study referred in the literature are the Pearson correlation coefficient (PLCC) and Spearman rank-order correlation coefficient (SROCC). Similarly, for the determination of the difference between the predicated score and the ground truth the famous metrics used are root mean square error (RMSE) and mean absolute error (MAE).

A number of BIQA approaches have been proposed in the past, however each of them have certain limitations. Most of these approaches treat the whole image as a unit and did take into account the fact that certain regions of an image may be more significant than others in determining overall image perceptual quality. This can lead to inaccuracies and reduced robustness in BIQA algorithms, particularly in cases where images contain multiple regions of interest, such as natural scenes or complex objects.

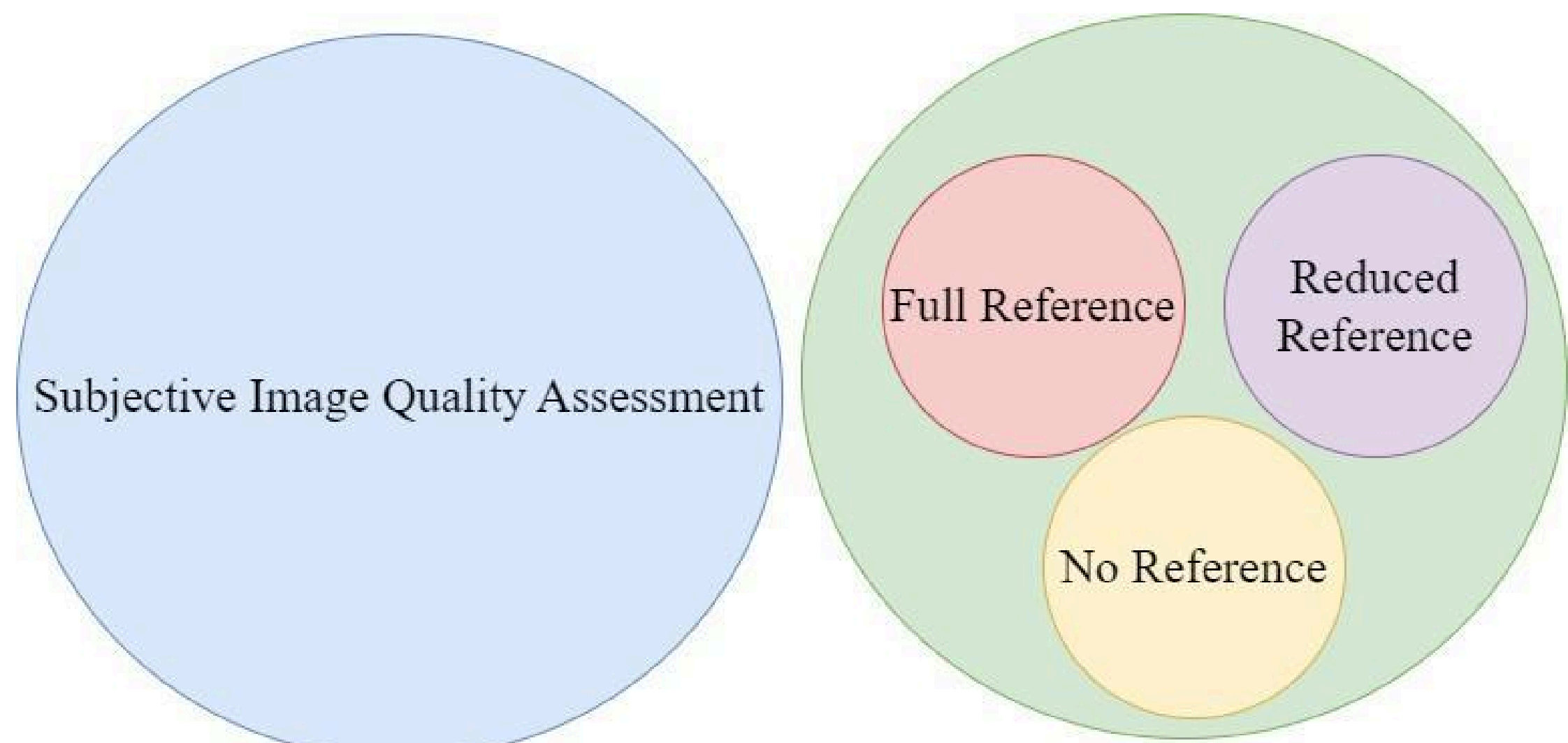

Figure 1: Relationship of various IQA techniques

To address the aforementioned limitations, a multi-stream spatial and channel attention algorithm has been proposed in this study. The algorithm generates more accurate predictions with a high correlation to human perceptual assessment by taking into account the hybrid feature fusion by two different backbones followed by spatial and channel attention for providing high weights to the region of interest. The demonstration of the spatial and channel attention on the object of interest is shown in the experiment's sections using GradCAM[26].

## 2 Literature Review

The literature review has been divided in to two parts. The first part discusses the hand-engineered features-based algorithms and the other part discusses the deep learning-based algorithms.

Moorthy *et.*al [22], propose a blind image quality assessment (IQA) algorithm referred to as DIIVINE. This algorithm is based on the hypothesis that pristine images exhibit certain statistical properties that are altered by artifacts and distortion, translating them into unnatural or unpristine images. Two stages are used in the proposed algorithm: the first stage identifies distortion; the second stage assesses distortion-specific image quality. Afterwards, Saad *et.*al [27] proposed a no-reference image quality assessment algorithm referred to as BLIINDS-II. The algorithm operates by assuming that images in their original form have specific statistical properties. However, when artifacts and distortions are introduced, these properties are altered resulting in images that appear unnatural or not pristine. To address this issue, the algorithm is designed to work in two stages. The first stage identifies the presence of any distortions while the second stage assesses the quality of the image based on the specific distortion identified. When the algorithm has been evaluated on the LIVE [28] IQA database, the algorithm has shown good correlation with the human provided subjective scores. Subsequently, Xue *et.*al [25] proposed a quality aware clustering method referred to as QAC for BIQA problem. The proposed algorithm has comparable performance in terms of linear correlation and accuracy as well. Then, Xue *et.*al [24], proposed a new model that uses two types of local contrast features - the gradient magnitude map and the Laplacian of Gaussian response. They tested this model on three large benchmark databases and compared it to other models that assess image quality, including some that use full reference images, and have reported the state-of-the-art performance. Zhang *et.*al[29], proposed a BIQA method that does not the human subjective opinion as the ground truth. Instead, the method utilizes a multivariate Gaussian model. The model is used with a set of patches of the images, that are collected from the pristine or natural images. The paper also discusses the natural image statistics and models used in the proposed method, including the Gaussian distribution of locally normalized luminances and the Weibull distribution of gradient magnitudes in natural images. The performance of the mode is quite reasonable. In the same year Fang et.al[8], devise a new way to solve BIQA (image quality assessment) problems pertaining to poor contrast images. The authors acknowledge that little research has been done in assessing the quality of contrast-distorted images in the absence of a perfect reference image. Based on several non-visual spectral shapes (NSS) features, they used support vector regression (SVR) to predict how humans would rate image quality, referred to as MOS. Later on, Xu et.al [30], proposed a method based on high order statistics aggregation (HOSA) for BIQA problem. It shows highly competitive performance to the state-of-the-art BIQA methods. The proposed method requires only a small codebook and has been evaluated on ten image databases with both simulated and realistic image distortions. The results show that incorporating high order statistics for image perceptual quality evaluation is effective and can solve the problems of previous feature learning-based BIQA methods. Ghadiyaram et.al[11], proposed proposes a "bag of feature maps" approach for blind image quality prediction that uses NSS to extract non-distortion specific features. The authors use the legacy database of authentically distorted and newly introduced distortion-realistic database referred to as LIVE In the Wild Image Quality Challenge Database [10]. The results have shown that the proposed algorithm has outperformed the existing algorithms. Kundu et.al[16], proposed a new NR IQA model for high dynamic range (HDR) images that is based on bandpass and differential NSS information of the HDR images. The proposed model uses both standard and newly designed HDR-space features as the part of the feature set. The proposed models are validated on both the HDR and standard images datasets and the performance of the model is optimal in both cases. Sadiq et.al [31], proposes a new blind image quality assessment (BIQA) technique that extracts features from both spatial and transform domains. The proposed technique is based on NSS and uses morphological gradient, discrete Laplacian, and stationary wavelet transform (SWT) to extract features for the images. The extracted features are then normalized using an adaptive joint normalization framework in order to avoid any type of redundancies. The proposed technique is evaluated on five legacy databases and has outperformed the state-of-the-art BIQA and full reference IQA techniques. Khalid et.al [5], proposed a gaussian process based BIQA algorithm referred to as GPR-BIQA. The author has proposed an integrated feature selection algorithm for the BIQA problem, a novel consolidation of features to propose a feature set incorporating the best features in transformed, spatial, and other domains, in the last a gaussian process-based regression algorithm has been proposed. The algorithm has been designed for non-distortion specific prediction of the quality of the images. The algorithm has been evaluated on natural and synthetically distorted databases. A comprehensive experimentation has been reported and it has been reported that the proposed algorithm has outperformed the existing NSS and Deep learning BIQA algorithms.

In addition to the NSS hand engineered based algorithms a number of deep learning-based algorithms have been proposed in the past with the state-of-the-art performance. Gu et.al [12], proposed a new Deep learning-based Image Quality assessment algorithm referred to as DIQI, which uses deep neural network based framework to capture the complex attributes of images. The DIQI is compared with classical full-reference and state-of-the-art reduced and no-reference IQA algorithms on the TID2013 [23] database and has outperformed them. Later on, Fu et.al [9], proposed a CNN based BIQA method. The algorithm has been evaluated on LIVE IQA database and the performance has been compared with the other algorithms. The algorithm has optimal performance. Subsequently, Bianco et.al [6], proposed a BIQA algorithm named as DeepBIQ. The algorithm has been compared with state-of-the-art algorithms on several legacy benchmark databases of synthetically distorted images, including LIVE[32], CSIQ[17], TID2008[33], and TID2013[23]. The results show that DeepBIQ outperforms the state-of-the-art methods. Then, Ma et.al [19], proposed a multi-task end-to-end optimized deep neural network referred to as MEON for BIQA problem. The model has been designed with two sub-networks, the first one is the distortion identification network, and the second one is the quality prediction network, it shares the early layers as well. The model has demonstrated strong competitiveness against state-of-the- art BIQA models when evaluated on IQA databases. Zhang et.al [34], in order to assess the quality of blind images, proposed a deep bilinear model that works for both synthetically and authentically distorted images. The model consists of two streams of deep convolutional neural networks (CNNs), specializing in two distortion scenarios separately. The algorithm has competitive performance. Ahmed et.al [1], proposed a CNN based BIQA algorithm. The author has proposed an ensemble technique and has reported the advantages of using the same for BIQA problem. Nisar et.al [3], proposed a hybrid method for BIQA problem. The proposed approach used the combination of hand-crafted and deep extracted features achieved higher performance as compared to the state-of-the-art algorithms in the same domain. The final model was tested on seven legacy benchmark databases for the comparative study. It has been reported that the proposed model has outperformed the existing model on the basis of correlation with the human opinion measure. Hosu et.al [13], reported a new dataset called KonIQ-10k, which is the largest claimed IQA dataset to date. They had also proposed a new deep learning-based model referred to as KonCept512, which had showed excellent generalization characteristics on the test set to the current state-of-the-art algorithm. Ahmed et.al [2], proposed a hybrid IQA model. The pre-trained deep learning architectures were experimented with, and their activations were used as features to train a Gaussian process regression model for quality assessment. The performance of the algorithm has been evaluated by using the correlation study. The method provided state-of-the-art performance and was validated on different legacy datasets. Zhang et.al [35], proposed a method for continual learning in blind image quality assessment (BIQA), where a model learns continually from a stream of IQA datasets, building on what was learned from previously seen data. The proposed method is shown to outperform standard training techniques for BIQA in extensive experiments. The paper also identifies five desiderata in the new setting with a measure to quantify the plasticity-stability trade-off. Ahmed et.al [4], proposed a model to evaluate the quality of digital images, a new measure called the Perceptual Image Quality Index (PIQI). In addition to calculating mean subtracted contrast normalized products in a variety of scales and color spaces, PIQI also computes luminance and gradient statistics. To conduct the perceptual quality assessment, these collected features are fed into a stacked ensemble of Gaussian Process Regression (GPR). Results from tests conducted on six benchmark datasets show that the PIQI performs competitively when compared against twelve state-of-the-art approaches. Ground truth and predicted quality ratings are compared using RMSE, Pearson, and Spearman's correlation coefficients and the algorithm has outperformed the existing algorithms.

# 3 Proposed Model

The end-to-end pictorial representation for the proposed architecture is reported in Figure 2 (a). The model consists of two ImageNet pretrained backbone followed by Global Average Pooling (GAP), spatial and channel attention, and a feed forward neural network in the end. The spatial and channel blocks are reported in Figure 2 (b) and (c). The spatial and channel blocks are inspired from MIRNET[36] algorithm. An image's spatial attention block is used to highlight the most important parts by learning weights for each pixel. By focusing on specific regions and suppressing irrelevant ones, the model is better able to recognize important features. On the other hand, a channel attention block is used to emphasize the important channels within a convolutional layer by learning weights that highlight the important channels and suppress the irrelevant ones. The model can then capture and utilize the most important information from each channel, resulting in a better overall feature representation.

The mathematical modelling for the proposed architecture is as below:

## 3.1 Spatial Block

The spatial attention block consists of four operations explained in details as below:

**a. Average Pooling:**

The average pooling operation calculates the average value along the last dimension of the input tensor x and results in an output tensor with the same shape except for the last dimension, which becomes 1. Mathematically, it can be expressed as:

$$\text{avg_pool}\,[i, j, k] = \frac{1}{\mathrm{H} \times \mathrm{W}} \sum_{h=1}^{\mathrm{H}} \sum_{w=1}^{\mathrm{W}} \mathrm{x}[i, h, w, k] \tag{1}$$

Where, $i$ represents the batch index, $j$ represents the height index, $k$ represents the channel index, H represents the height of the input tensor, and W represents the width of the input tensor.

**b. Maximum Pooling:**

The maximum pooling operation calculates the maximum value along the last dimension of the input tensor x and results in an output tensor with the same shape except for the last dimension, which becomes 1. Mathematically, it can be expressed as:

$$\text{max_pool}\,[i, j, k] = \max_{h=1}^{H} \max_{w=1}^{W} x[i, h, w, k] \tag{2}$$

where the indices have the same meaning as described in the previous equation.

**c. Concatenation:**

The concatenation operation concatenates the output tensors from the average pooling and maximum pooling operations along the last dimension, resulting in an output tensor with a larger last dimension. Mathematically, it can be expressed as:

$$\text{concat}\,[i, j, k] = [\,\text{avg_pool}\,[i, j, k], \text{maxpool}\,[i, j, k]] \tag{3}$$

**d. Convolution:**

The convolution operation applies a 2D convolution with a 3x3 kernel to the concatenated tensor concat, producing an output tensor of shape (batch_size, height, width,1) with the sigmoid activation applied element-wise. Mathematically, it can be expressed as:

$$\text{conv}\,[i, j, k] = \sigma\left(\sum_{m=-1}^{1} \sum_{n=-1}^{1} w[m, n] \cdot \text{concat}\,[i, j + m, k + n] + b\right) \tag{4}$$

where, w represents the 3x3 convolution kernel, σ represents the sigmoid activation function, b represents the bias term, and the indices have the same meaning as described in the previous equations.

**e. Element-wise Multiplication:**

The element-wise multiplication operation performs element-wise multiplication between the input tensor x and the output tensor conv obtained from the convolution operation. This operation produces the final output tensor. Mathematically, it can be expressed as:

$$\text{output}\,[i, j, k] = x[i, j, k] \cdot \text{conv}\,[i, j, k] \tag{5}$$

where the indices have the same meaning as described in the previous equations.

Note: In the above equations, σ represents the sigmoid activation function, and w and b are the parameters of the convolution operation. The symbol · represents element-wise multiplication between two tensors of the same shape.

### 3.2 Channel Attention Block

The channel attention layer takes an input tensor x and performs the following operations:

**a. Average Pooling:**

The average pooling operation calculates the average value along the spatial dimensions (height and width) of the input tensor x and results in an output tensor with the shape (batch_size,1,1, channels). Mathematically, it can be expressed as:

$$\text{avg_pool}\,[i,1,1,k] = \frac{1}{H \times W} \sum_{h=1}^{H}\sum_{w=1}^{W} x[i,h,w,k] \quad (6)$$

**b. Maximum Pooling:**

The maximum pooling operation calculates the maximum value along the spatial dimensions (height and width) of the input tensor x and results in an output tensor with the shape (batch_size,1,1, channels). Mathematically, it can be expressed as:

$$\text{max_pool}\,[i,1,1,k] = \max_{h=1}^{H}\max_{w=1}^{W} x[i,h,w,k] \quad (7)$$

**c. Concatenation:**

The concatenation operation concatenates the output tensors from the average pooling and maximum pooling operations along the last dimension, resulting in an output tensor with the shape (batch_size,1,1,2×channels). Mathematically, it can be expressed as:

$$\text{concat}[i,1,1,k] = [\,\text{avg_pool}\,[i,1,1,k], \text{max_pool}\,[i,1,1,k]] \quad (8)$$

where the indices have the same meaning as described in the previous equations.

**d. Fully Connected Layers:**

The two fully connected layers are applied consecutively to perform two operations on the concatenated tensor, referred to as "concat" in equation (8).

The first fully connected layer, referred as dense1, reduces the number of channels by a factor of 8 and applies the ReLU activation function. Mathematically, it can be expressed as:

$$\text{dense 1}[i,1,1,k] = \text{ReLU}\,\Big(\sum_{j=1}^{2\times\text{ channels}} W_{\text{densel}}\,[j,k]\cdot \text{concat}\,[i,1,1,j] + b_{\text{densel}}\,[k]\Big) \quad (9)$$

where $W_{\text{densel}}$ represents the weight matrix and $b_{\text{densel}}$ represents the bias term of the first fully connected layer.

The second fully connected layer, referred to as dense2, restores the original number of channels and applies the sigmoid activation function. Mathematically, it can be expressed as:

$$\text{dense2}[i,1,1,k] = \text{Sigmoid}\left(\sum_{j=1}^{\text{channels}} W_{\text{dense2}}[j,k] \cdot \text{dense1}[i,1,1,j] + b_{\text{dense2}}[k]\right) \quad (10)$$

where $W_{\text{densel}}$ represents the weight matrix and $b_{\text{densel}}$ represents the bias term of the first fully connected layer.

**e. Element-wise Multiplication:**

The element-wise multiplication operation performs element-wise multiplication between the input tensor x and the output tensor dense2 obtained from the fully connected layers. This operation produces the final output tensor. Mathematically, it can be expressed as:

$$\text{output}[i,h,w,k] = x[i,h,w,k] \cdot \text{dense 2}[i,1,1,k] \quad (11)$$

Where, the indices have the same meaning as described in the previous equations. Note: In the above equations, ReLU represents the rectified linear unit activation function, Sigmoid represents the sigmoid activation function, and $W_{\text{densel}}$ , $b_{\text{densel}}$ , $W_{\text{dense 2}}$, $b_{\text{dense2}}$ are the parameters (weights and biases) of the fully connected layers.

### 3.3 Final Proposed Model

We define the complete model architecture as follows:

**a. Input layer:**

The input tensor is denoted as x.

**b. ResNet50 Backbone:**

The ResNet50 backbone takes the input tensor x and performs the following operations:

$$x_2 = \text{ResNet 50}(x) \quad (12)$$

$$x_2 = \text{GlobalAveragePooling2 D}(x_2) \quad (13)$$

$$x_2 = \text{Reshape}((1,1,-1))(x_2) \quad (14)$$

$$x_2 = \text{spatial_attention}(x_2) \quad (15)$$

$$x_2 = \text{channel_attention}(x_2) \quad (16)$$

$$x_2 = \text{Flatten}(x_2) \quad (17)$$

**c. EfficientNetB7 Backbone:**

The EfficientNetB7 backbone takes the input tensor x and performs the following operations:

$$x_3 = \text{EfficientNetB7}\ (x) \tag{18}$$

$$x_3 = \text{GlobalAveragePooling 2D}(x_3) \tag{19}$$

$$x_3 = \text{Reshape}\ ((1,1,-1))(x_3) \tag{20}$$

$$x_3 = \text{spatial_attention}\ (x_3) \tag{21}$$

$$x_3 = \text{channel attention}\ (x_3) \tag{22}$$

$$x_3 = \text{Flatten}\ (x_3) \tag{23}$$

**d. Concatenated Output:**

The outputs from the ResNet50 and EfficientNetB7 backbones are concatenated to form a single tensor:

$$\text{concatenated_output} = \text{Concatenate}\ ()([x_2, x_3]) \tag{24}$$

**e. Fully Connected Layers:**

The concatenated output tensor mentioned in equation (24) afterwards undergoes several fully connected layers with batch normalization and dropout applied:

$$x = \text{Dense}\ (1024, \text{activation} =' \text{relu}\ ')(\ \text{concatenated_output}) \tag{25}$$

$$x = \text{BatchNormalization}\ (x) \tag{26}$$

$$x = \text{Dropout}\ (0.25)(x) \tag{27}$$

$$x = \text{Dense}\ (512, \text{activation} =' \text{relu}')(x) \tag{28}$$

$$x = \text{BatchNormalization}\ (x) \tag{29}$$

$$x = \text{Dropout}\ (0.25)(x) \tag{30}$$

$$x = \text{Dense}\ (256, \text{activation} =' \text{relu}')(x) \tag{31}$$

$$x = \text{BatchNormalization}\ (x) \tag{32}$$

$$x = \text{Dropout}\ (0.5)(x) \tag{33}$$

### f. Regression Head:

The final stage is referred to as regression head. It predicts the final output using a dense layer with linear activation:

$$\text{predictions} = \text{Dense}\ (\text{num_classes, activation} = '\text{linear}')(x) \tag{24}$$

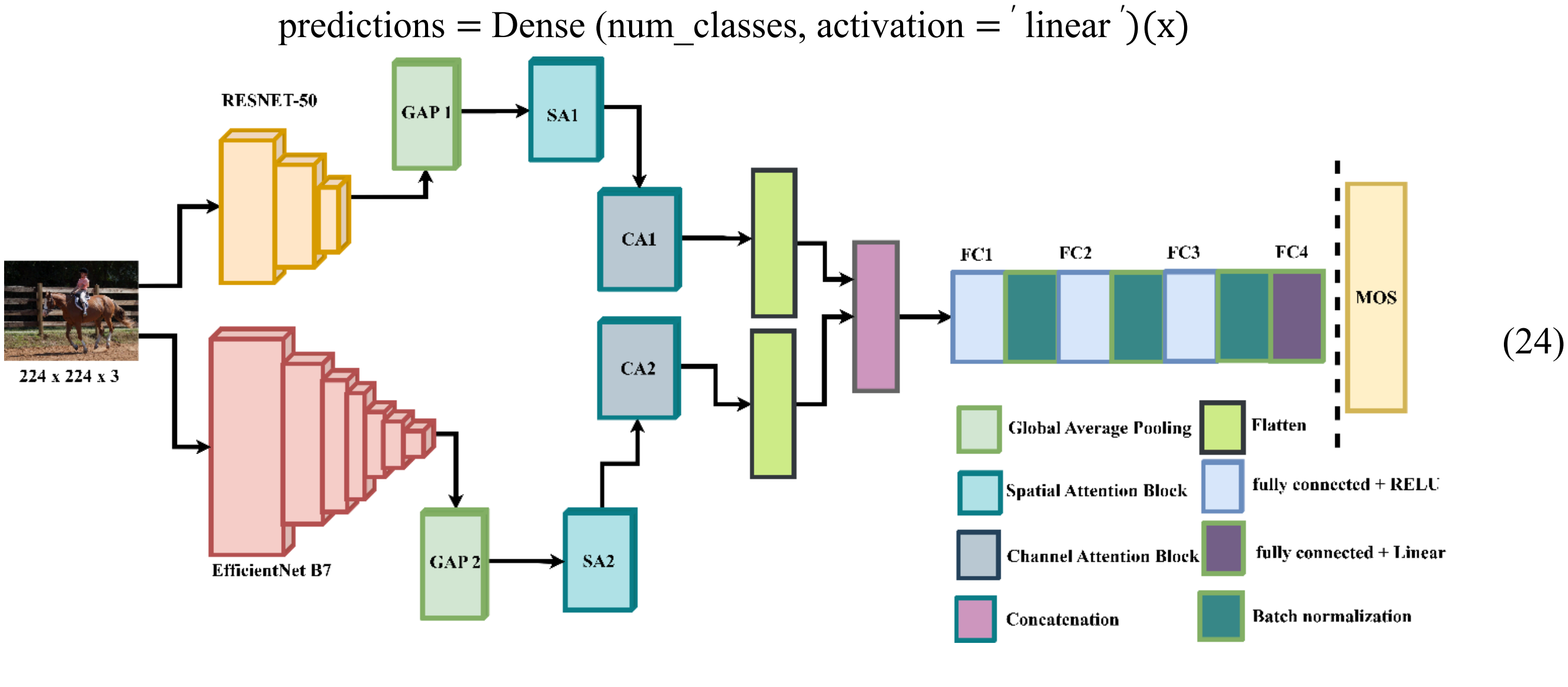

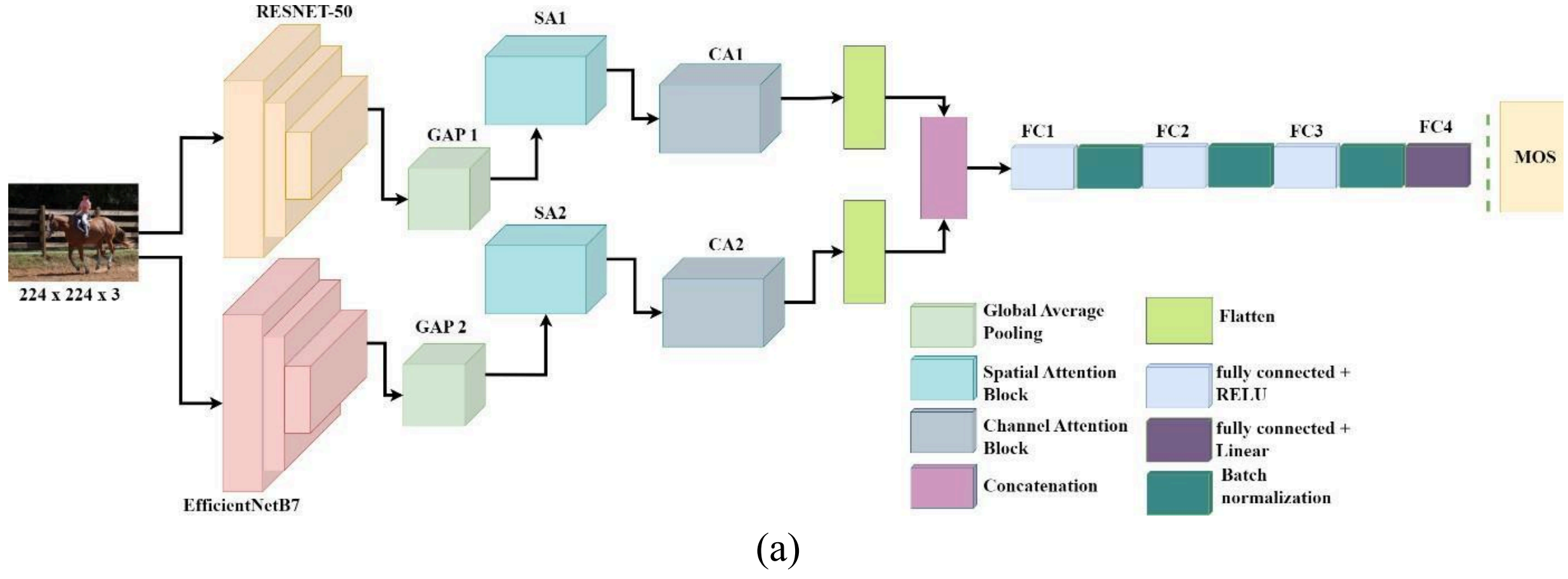

(a)

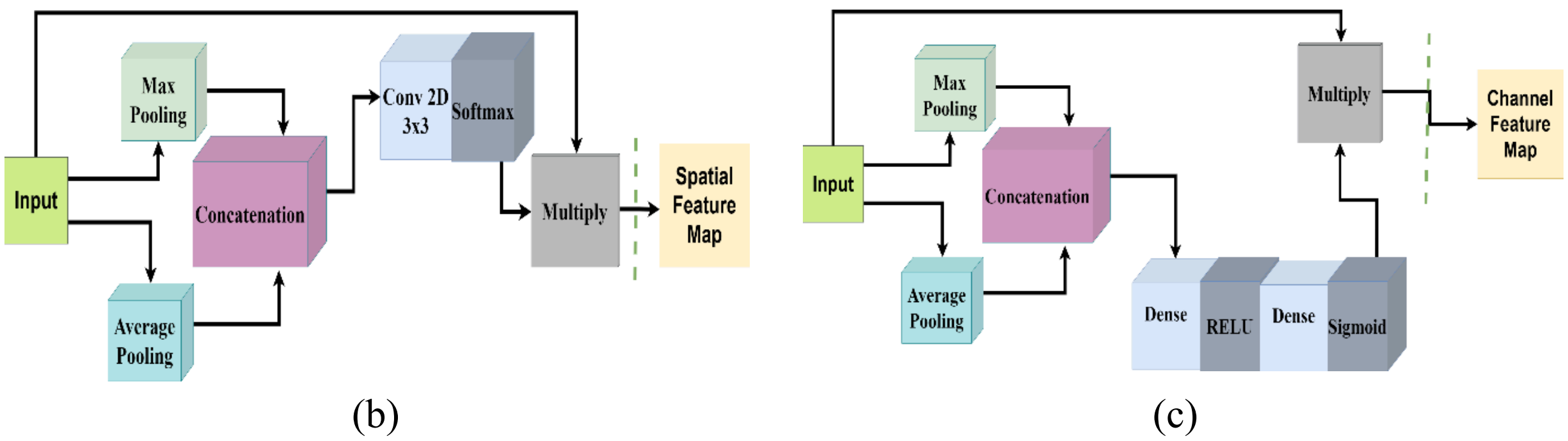

Figure 2: (a) Architecture of the end-to-end system;(b) Architecture for Spatial attention block;(c) Architecture for channel attention block

# 4 Experimental Results

## 4.1 Datasets

For evaluating the performance of BIQA models, the choice of dataset is very important. Therefore, in order to conduct an in-depth and comprehensive evaluation, we have utilized both synthetic and natural distortions image datasets. Using the synthetic distortions image datasets, the performance of BIQA models under controlled or laboratory conditions has been evaluated initially using the synthetic image dataset, whereas the performance of models for the real-world conditions has been evaluated using natural image datasets.

We have utilized the TID2013 dataset, which contains 25 reference images and 24 types of different distortions at five distinct distortion levels. The CSIQ dataset, which is comprised of 30 reference images and six types of distortions with at maximum five distinct distortion levels. The LIVE dataset, which is comprised of 29 reference images and five types of distortion at five distinct levels of distortion. The LIVE in the wild Challenge dataset, which consists of 1161 images with varying degrees and types of distortion. The images included in this dataset are captured by different and diverse image capturing devices, so that the natural artifacts and distortions may be authentically captured. Finally, we have utilized the Koniq-10k dataset, which consists of 10073 naturally distorted images, the dataset is one of the most diverse and large-scale datasets available for the design and evaluation of the image quality assessment algorithm in the natural environment. The summary for the datasets is tabulated in Table.1.

Table 1: Summary of the used Image Quality Assessment Datasets

| Dataset | Type | Total Images | Resolution | Score Range |
|---|---|---|---|---|
| Koniq-10k | Authentically distorted | 10073 | Multiple | 0-100 |
| BIQ2021 | Authentically distorted | 8000 | 512x512 | 0-1 |
| LIVEC | Authentically distorted | 1162 | 500x500 | 0-100 |
| TID2013 | Synthetically distorted | 3000 | 512x384 | 0-9 |
| CSIQ | Synthetically distorted | 866 | 512x512 | 0-100 |

## 4.2 Evaluation Metrics

For the validation and comparison of the performance of the designed algorithm the standard evaluation metrics used for the typical regression problem are utilized. The metrics used are as below:

1. Pearson Linear Correlation Coefficient (PLCC)
2. Spearman Ranked Order Correlation Coefficient (SROCC)
3. Kendall Ranked Order Correlation Coefficient (KROCC)

The details for each of the performance evaluation metric is discussed in the relevant section.

### 4.2.1 Pearson Linear Correlation Coefficient (PLCC)

The Pearson correlation coefficient is widely used to quantify the strength of a linear relationship. In the context of image quality assessment, it is commonly used to evaluate how well a BIQA model's predicted quality scores match up with the ground truth quality scores. The formula for PLCC is:

$$PLCC = \frac{\mathrm{Cov}\,(Y, \hat{Y})}{\sqrt{\mathrm{Var}\,(Y)\mathrm{Var}\,(\hat{Y})}}$$

The value of PLCC ranges from -1 to +1. A PLCC value of +1 indicates a perfect positive linear correlation, meaning that as the ground truth quality scores increase, the BIQA model's predicted scores also increase proportionally. A PLCC value of -1 indicates a perfect negative linear correlation, meaning that as the ground truth quality scores increase, the BIQA model's predicted scores decrease proportionally. Finally, a PLCC value of 0 indicates that there is no linear correlation between the two sets of scores.

### 4.2.2 Spearman Ranked Order Correlation Coefficient (SROCC)

The Spearman Rank Correlation Coefficient (SROCC) is a tool for quantifying the strength of a monotonic relationship between two variables. It is used in image quality assessment to compare a model's predicted quality scores with the ground truth quality scores, regardless of the exact magnitude of the scores. The formula for SROCC is:

$$SROCC = 1 - \frac{6\sum D^2}{n(n^2 - 1)}$$

where D is the difference between the ranks of Y and $\hat{Y}$, and n is the number of samples.

The value of SROCC ranges between -1 and +1. A SROCC value of +1 indicates a perfect monotonic correlation, meaning that as the ground truth quality scores increase (or decrease), the BIQA model's predicted scores also increase (or decrease). A SROCC value of -1 indicates a perfect negative monotonic correlation, meaning that as the ground truth quality scores increase (or decrease), the BIQA model's predicted scores decrease (or increase). Finally, a SROCC value of 0 indicates no monotonic correlation between the two sets of scores.

### 4.2.3 Kendall Ranked Order Correlation Coefficient (KROCC)

The Kendall tau is another measure of the correlation between the predicted and ground truth variables. It is specifically, used to measures the similarity between the rankings of the two variables.

The formula for calculating Kendall tau is:

$$\tau = \frac{2P}{n(n-1)}$$

where, P is the number of concordant pairs minus the number of discordant pairs, and n is the number of observations. A concordant pair is a pair of observations that have the same order in both variables being compared (i.e., they are either both greater than or both less than each other). A discordant pair is a pair of observations that have opposite orders in the two variables (i.e., one is greater than the other in one variable, but smaller in the other variable).

The Kendall tau coefficient (denoted by the symbol τ) ranges from -1 to +1, with values of -1 indicating a perfect negative association, 0 indicating no association, and +1 indicating a perfect positive association.

## 4.3 Implementation details

The proposed algorithm has been trained and evaluated using the NVIDIA P5000 graphics card, which has a 16 GB memory bandwidth. Adam is the optimizer used to train the algorithm. The training learning rate begins at 1e-4 and decreases if the training progress (as determined by validation) does not improve for two consecutive epochs. The minimum learning rate specified is 1e-8. In addition to preventing overfitting, the early stopping criterion has been imposed. The total number of training epochs is 100, and the batch size is 10. The model's image dimensions are 224 x 224,

Each of the dataset is divided in to three non-overlapping splits referred to as training, validation, and testing. The splits are than used for the training, testing of the proposed model and later on for the comparison with the existing best performing models. For the fair comparison each of the algorithm is evaluated on the same image size and split.

## 4.4 Loss Function

A modified loss function has been proposed for the training of the algorithm to catch better correlation and absolute value. The loss function is reported below:

$$\text{Loss} = \lambda_1 * \text{MAE} + \lambda_2 * \text{PLCC Loss}$$

Where;

$$MAE = \frac{1}{n}\sum_{i=1}^{n} |Y_i - \hat{Y}_i|$$

And

$$\text{PLCC Loss} = 1 - \frac{\text{Cov}(Y, \hat{Y})}{\sqrt{\text{Var}(Y)\text{Var}(\hat{Y})}}$$

where $Y_i$ is the ground truth score, and $\hat{Y}_i$ is the predicted score, and n is the number of samples. Here, $\lambda_1=1$ and $\lambda_2=10$. The weight terms assigned to each of the component of the loss function have been finalized by using parametric analysis. The significance of using a combined MAE and
PLCC loss function for regression tasks lies in its robustness to outliers, assessment of linear relationships, balanced evaluation of accuracy and correlation, customization of performance based on task requirements, and potential enhancement of model generalizability.

## 4.5 Performance Evaluation

Blind Image Quality Assessment (BIQA) is an important area of research that seeks to automatically assess the quality of an image without knowing the original image. Therefore, we use qualitative and quantitative analysis techniques to improve the reliability of our BIQA algorithms.

### 4.5.1 Quantitative analysis

Assessment (BIQA), as they provide a rigorous and objective method of evaluating image quality algorithms. Image quality metrics such as Pearson Linear Correlation Coefficient (PLCC), and Spearman Rank Order Correlation Coefficient (SROCC), and Kendall rank correlation coefficient (KRCC) are usually used to measure quantitative information using mathematical and statistical techniques. These quantitative methods are useful for assessing image quality on a large scale.

The quantitative analysis has been divided into two parts: they are focused on training the algorithm on three naturally distorted and one synthetic distortion datasets, which are then compared with other algorithms in a comparative analysis. Although there are a number of other datasets also available, however the performance has been evaluated and compared on the largest datasets.

Table 2 summarizes the findings for the TID2013 dataset. Here, it can be noted that among all the compared algorithms the proposed approach has outperformed optimally. However, the distortions and artifacts introduced into images in real time can be complex, which means the model trained on the synthetic distortion datasets can't accurately predict the quality, so it has been tested and evaluated on the largest natural or authentic distortion datasets. Table 3,4 and 5 summarizes the findings for the KONIQ-10k, LIVEC, and BIQ2021 datasets. Overall, we can conclude that the proposed method out performs the best among the evaluated metrics in terms of PLCC, SRCC, and KRCC.

The PLCC, SROCC, and KRCC are the primary comparison elements. As in the BIQA domain, the most significant concern is the correlation between predicted and true ground-truth variables, as the primary goal is to mimic human opinion. The top three performing algorithms are compared on the basis of the highest SRCC, PLCC and KRCC values. The blue color represents the first in rank, green represents the second, and red represents the third.

Table 2: The PLCC, SRCC and KRCC values of various methods on TID2013 dataset. Blue, green and red refers to the best, second and third score among all comparison methods, respectively.

| *Algorithm Name* | **Performance Metrics** | | |
|---|---|---|---|
| | **PLCC** | **SRCC** | **KRCC** |
| NIQE*[21]* | 0.37 | 0.31 | 0.21 |
| CNNIQA*[14]* | 0.39 | 0.16 | 0.11 |
| BRISQUE*[20]* | 0.43 | 0.37 | 0.26 |
| PI*[7]* | 0.45 | 0.34 | 0.24 |
| NRQM*[18]* | 0.46 | 0.33 | 0.23 |

| ILNIQE*[29]* | 0.52 | 0.49 | 0.34 |
|---|---|---|---|
| DBCNN*[34]* | 0.55 | 0.43 | 0.31 |
| PAQ2PIQ*[37]* | 0.58 | 0.40 | 0.28 |
| CLIPIQA+_VITL14_512*[38]* | 0.61 | 0.53 | 0.37 |
| CLIPIQA*[38]* | 0.65 | 0.58 | 0.41 |
| MUSIQ-KONIQ*[15]* | 0.68 | 0.58 | 0.41 |
| CLIPIQA+_RN50_512*[38]* | 0.69 | 0.59 | 0.42 |
| MANIQA*[39]* | 0.69 | 0.59 | 0.42 |
| CLIPIQA+*[38]* | 0.70 | 0.63 | 0.45 |
| GPR-BIQA*[5]* | 0.91 | 0.90 | 0.73 |
| **Proposed Method** | 0.92 | 0.93 | 0.77 |

Table 3: The PLCC, SRCC and KRCC values of various methods on KONIQ-10k dataset. Blue, green and red refers to the best, second and third score among all comparison methods, respectively.

| ***Algorithm Name*** | **Performance Metrics** | | |
|---|---|---|---|
| | **PLCC** | **SRCC** | **KRCC** |
| BRISQUE*[20]* | 0.21 | 0.23 | 0.15 |
| NRQM*[18]* | 0.48 | 0.37 | 0.25 |
| NIQE*[21]* | 0.32 | 0.38 | 0.26 |
| PI*[7]* | 0.47 | 0.46 | 0.31 |
| ILNIQE*[29]* | 0.52 | 0.55 | 0.39 |
| PAQ2PIQ*[37]* | 0.71 | 0.64 | 0.46 |
| MANIQA*[39]* | 0.72 | 0.66 | 0.47 |
| CLIPIQA*[38]* | 0.72 | 0.66 | 0.47 |
| GPR-BIQA*[5]* | 0.71 | 0.68 | 0.52 |
| CNNIQA*[14]* | 0.80 | 0.76 | 0.56 |
| CLIPIQA+*[38]* | 0.85 | 0.80 | 0.61 |
| DBCNN*[34]* | 0.86 | 0.84 | 0.66 |
| CLIPIQA+_VITL14_512*[38]* | 0.87 | 0.84 | 0.67 |
| MUSIQ-KONIQ*[15]* | 0.85 | 0.84 | 0.65 |
| CLIPIQA+_RN50_512*[38]* | 0.86 | 0.84 | 0.67 |
| **Proposed Method** | 0.86 | 0.88 | 0.67 |

Table 4: The PLCC, SRCC and KRCC values of various methods on LIVEC dataset. Blue, green

and red refers to the best, second and third score among all comparison methods, respectively.

| *Algorithm Name* | **Performance Metrics** | | |
|---|---|---|---|
| | **PLCC** | **SRCC** | **KRCC** |
| NRQM*[18]* | 0.41 | 0.30 | 0.20 |
| BRISQUE*[20]* | 0.35 | 0.31 | 0.21 |
| ILNIQE*[29]* | 0.49 | 0.44 | 0.30 |
| NIQE*[21]* | 0.48 | 0.45 | 0.31 |
| PI*[7]* | 0.52 | 0.46 | 0.31 |
| CNNIQA*[14]* | 0.63 | 0.61 | 0.43 |
| MANIQA*[39]* | 0.72 | 0.66 | 0.47 |
| GPR-BIQA*[5]* | 0.66 | 0.66 | 0.46 |
| CLIPIQA*[38]* | 0.69 | 0.70 | 0.51 |
| PAQ2PIQ*[37]* | 0.75 | 0.72 | 0.53 |
| DBCNN*[34]* | 0.79 | 0.76 | 0.57 |
| CLIPIQA+_VITL14_512*[38]* | 0.77 | 0.77 | 0.57 |
| MUSIQ-KONIQ*[15]* | 0.83 | 0.79 | 0.60 |
| CLIPIQA+*[38]* | 0.83 | 0.80 | 0.61 |
| CLIPIQA+_RN50_512*[38]* | 0.82 | 0.82 | 0.62 |
| **Proposed Method** | 0.86 | 0.84 | 0.68 |

Table 5: The PLCC, SRCC and KRCC values of various methods on BIQA2021 dataset. Blue, green and red refers to the best, second and third score among all comparison methods, respectively.

| *Algorithm Name* | **Performance Metrics** | | |
|---|---|---|---|
| | **PLCC** | **SRCC** | **KRCC** |
| ILNIQE*[29]* | 0.28 | 0.26 | 0.20 |
| NIQE*[21]* | 0.30 | 0.27 | 0.31 |
| NRQM*[18]* | 0.42 | 0.30 | 0.21 |
| PI*[7]* | 0.53 | 0.46 | 0.32 |
| BRISQUE*[20]* | 0.70 | 0.60 | 0.31 |
| CNNIQA*[14]* | 0.64 | 0.61 | 0.43 |
| MANIQA*[39]* | 0.73 | 0.66 | 0.48 |
| GPR-BIQA*[5]* | 0.66 | 0.66 | 0.46 |
| CLIPIQA*[38]* | 0.69 | 0.70 | 0.51 |
| PAQ2PIQ*[37]* | 0.76 | 0.72 | 0.54 |
| DBCNN*[34]* | 0.79 | 0.76 | 0.57 |

| CLIPIQA+_VITL14_512*[38]* | 0.77 | 0.77 | 0.58 |
|---|---|---|---|
| MUSIQ-KONIQ*[15]* | 0.83 | 0.79 | 0.60 |
| CLIPIQA+*[38]* | 0.84 | 0.81 | 0.62 |
| CLIPIQA+_RN50_512*[38]* | 0.82 | 0.82 | 0.62 |
| **Proposed Method** | 0.85 | 0.86 | 0.67 |

### 4.5.2 Qualitative analysis

#### *4.5.2.1 Regression Analysis*

This part of the evaluation involves fitting the best line to the data to determine the relationship between the target and the prediction. Fitting a line to the data that best represents the relationship between the predicted and actual MOS values. By examining the slope, intercept, and R-squared value of the line, we can learn about the accuracy and precision of the algorithm's predictions. The results for the same for the two largest datasets are reported in Figure 2. The best fit line and target line are very close to each other here. The predicted values are showing high correlation with the actual or target line. The study was designed to design an algorithm that would have a high correlation to humans, and therefore, it succeeded in that goal.

#### *4.5.2.2 Distribution Analysis*

In the regression analysis the distribution of the predicted variable is of an important concern as the same has to be closely related to the distribution of ground truth. Therefore, the analysis has been performed to examine the distributions of the both variables. The analysis has been performed for the two biggest authentically distorted datasets as in the case of regression Analysis. The results are reported in Figure 2. Here, it can be noted that the predication variable has the density with almost the same mean, standard deviation, and type as that of the ground-truth variables. Therefore, it can be deduced that the proposed algorithm is emulating the human perceptual behavior in terms of image quality assessment.

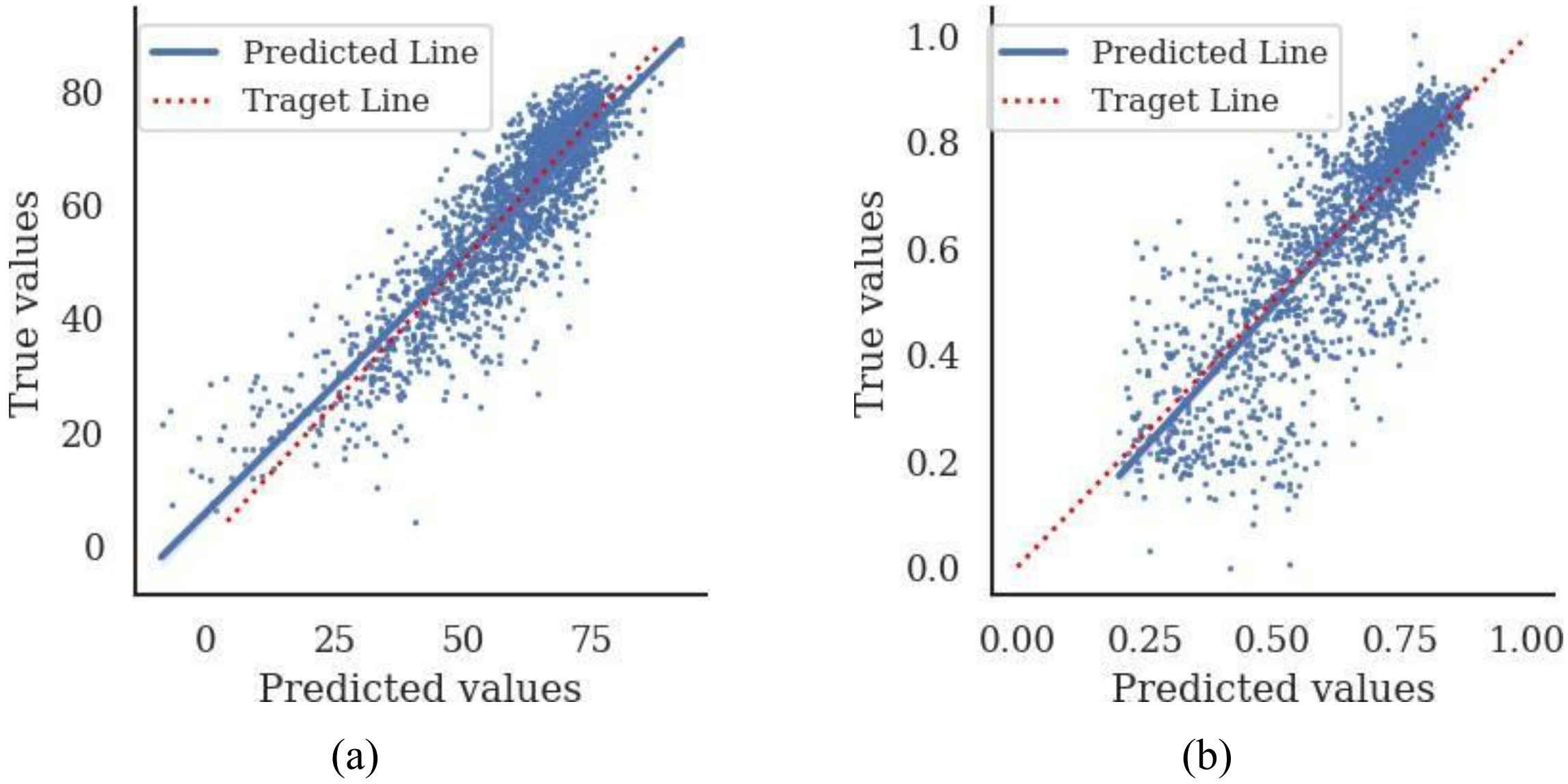

Figure 1: Scatter-plot between ground-truth versus predicted values along with regression line for (a) KONQ-10k (b) BIQ2021

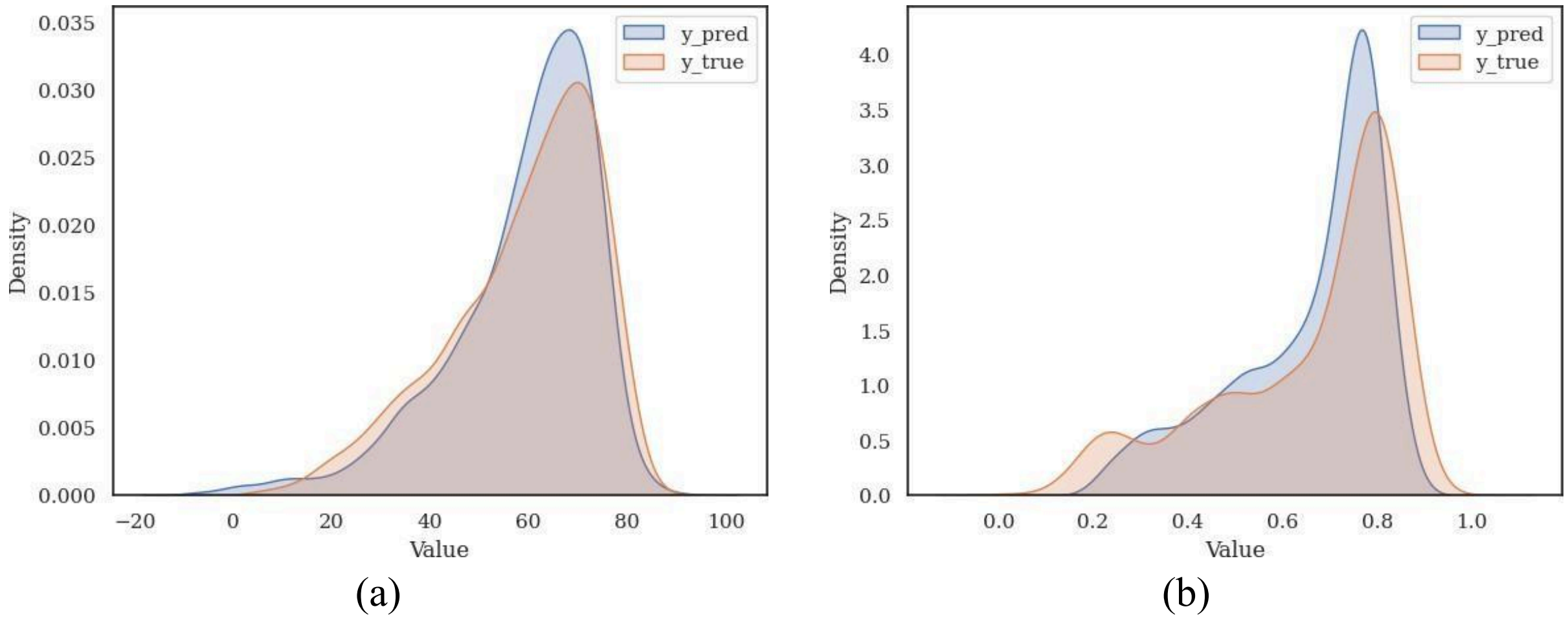

Figure 2: Distribution plot between ground-truth versus predicted values along for (a) KONQ-10k (b) BIQA2021

### 4.6 Ablation study

Ablation studies in deep learning allow for a better understanding of the relationships between parameters within a network. By selectively removing certain parts of the network, we can determine the most effective components in terms of performance and accuracy. The ablation study has been performed on the proposed algorithm and the results for the same are reported in Figure
2. The experimentation has been performed by using the official split for KONIQ-10k dataset.

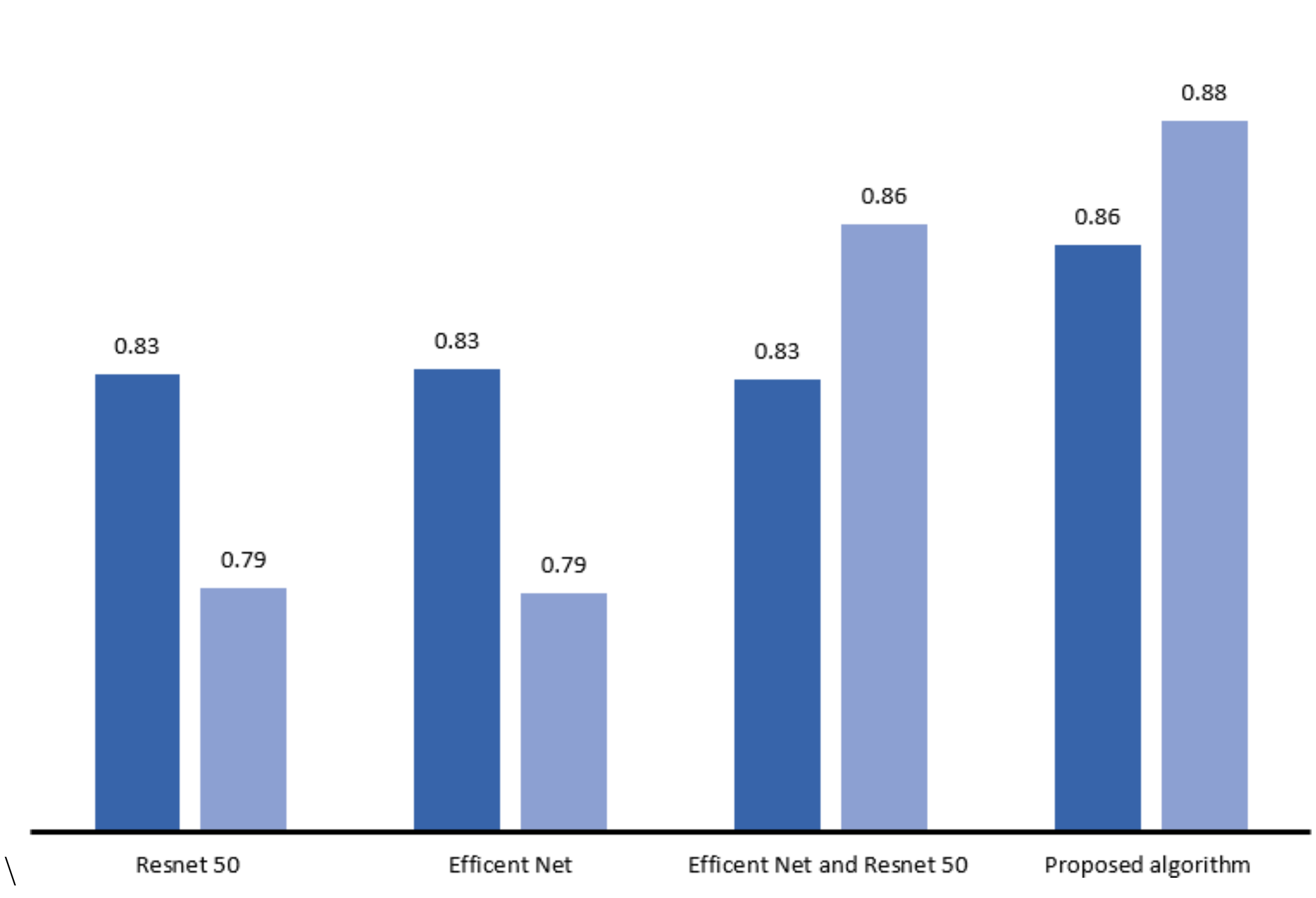

Figure 2: Ablation study

### 4.7 GRADCAM based visualization

We applied GRADCAM to a random set of images to understand the impact of attention and effectiveness of the proposed architecture. It can be observed here that the proposed method gives greater weight to foreground information than background. Furthermore, Figure 4 (b), and (d) indicate that the algorithm focuses primarily on information that directly affects perception, since humans can just click on a text to understand the distortion, and likewise on faces and eyes. The proposed algorithm provides optimal attention maps and is efficient and generalizable based on the results.

The proposed algorithm attention maps seem to do their jobs optimally, and its efficiency with good generalizability is confirmed by the results.

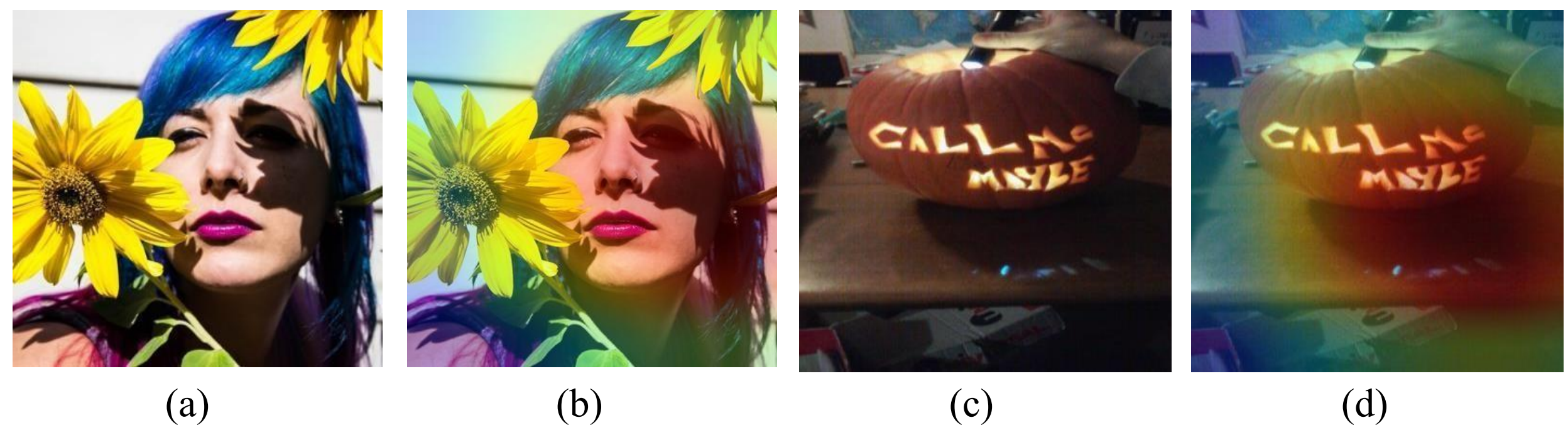

(a) (b) (c) (d)

Figure 4: GRADCAM based heatmaps

## Conclusion

This study proposes a multiteam fusion network with spatial and channel attention. The proposed architecture has been quantitatively and qualitatively evaluated, and it is compared to the domain's well-known best-performing algorithms. An in-depth investigation revealed that the proposed algorithm outperformed the existing algorithm in terms of all the famous correlation coefficients on the legacy largest natural and synthetic datasets. Furthermore, the spatial and channel attention mechanism and the algorithm's effectiveness have been demonstrated using attention visualization methodology. The proposed algorithm has been observed to focus on critical areas of the image and make decisions based on effective information. Because the algorithm has a high correlation with human opinion, it can be used as an excellent candidate for blind image quality assessment.